\def\year{2020}\relax
\documentclass[letterpaper]{article} 
\usepackage{aaai20}  
\usepackage{times}  
\usepackage{helvet} 
\usepackage{courier}  
\usepackage[hyphens]{url}  
\usepackage{graphicx} 
\usepackage{epsfig}
\usepackage{amsmath}
\usepackage{amssymb}
\usepackage{subcaption}
\usepackage{makecell}
\usepackage{url}
\usepackage{xcolor}
\usepackage{colortbl}
\usepackage{verbatim}

\usepackage{graphicx}  
\title{Assembling Semantically-Disentangled Representations for Predictive-Generative Models via Adaptation from Synthetic Domain}
\author{
\Large \textbf{Burkay Donderici}\\
Mythic Inc. \\ 
333 Twin Dolphin Dr STE 300\\
Redwood City, CA 94065\\
burkay.donderici@mythic-ai.com\\ 
\And
\Large \textbf{Caleb New}\\
Freeus, LLC.\\
640 W 1100 S\\
Ogden, UT 84404\\
caleb.new@freeus.com\\
\\
\And
\Large \textbf{Chenliang Xu}\\ 
University of Rochester\\
3005 Wegmans Hall, Hutchison Road,\\
University of Rochester, Rochester, NY 14627\\
chenliang.xu@rochester.edu
}
 
\begin{document}

\maketitle

\begin{abstract}
   Deep neural networks can form high-level hierarchical representations of input data. Various researchers have demonstrated that these representations can be used to enable a variety of useful applications. However, such representations are typically based on the statistics within the data, and may not conform with the semantic representation that may be necessitated by the application. Conditional models are typically used to overcome this challenge, but they require large annotated datasets which are difficult to come by and costly to create. In this paper, we show that semantically-aligned representations can be generated instead with the help of a physics-based engine. This is accomplished by creating a synthetic dataset with decoupled attributes, learning an encoder for the synthetic dataset, and augmenting prescribed attributes from the synthetic domain with attributes from the real domain. It is shown that the proposed (SYNTH-VAE-GAN) method can construct a conditional predictive-generative model of human face attributes without relying on real data labels.
\end{abstract}

\section{Introduction}


Deep neural networks are well suited to map complicated relationships between pairs of inputs and outputs. This has been exploited in computer vision tasks that involve image classification, transformation or generation to great success~\cite{krizhevsky2012imagenet,sermanet2013overfeat,simonyan2014very,ioffe2015batch,szegedy2017inception}. This extraordinary ability of deep networks is achieved as a result of construction of higher-level representations of inputs within the network, either implicitly as in classification tasks, or explicitly as in generation tasks~\cite{goodfellow2014generative,yosinski2015understanding,wu2016learning}. For example, in~\cite{radford2015unsupervised}, arithmetic operations in the latent representation space are shown to result in semantic operations in the outputs. 

\begin{figure}[t]
    \centering
    \includegraphics[width=2.3in,height=1.7in]{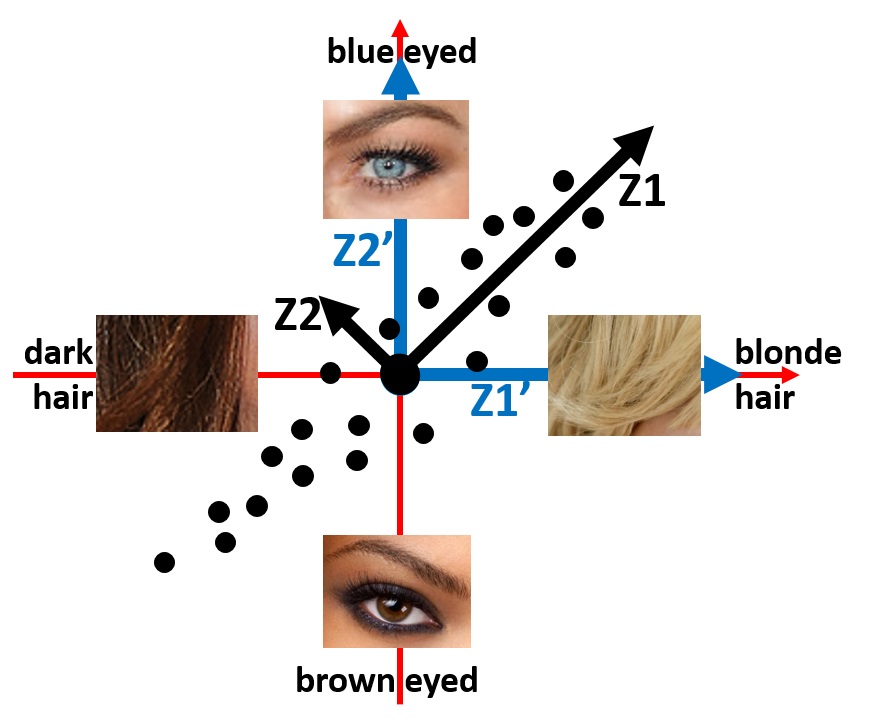}
	\caption{Illustration contrasting statistical disentanglement with semantic disentanglement. Black/Blue vectors indicate the coordinates of the real data embedding based on stati
	stical/semantic disentanglement, respectively. The goal of SYNTH-VAE-GAN is to achieve the semantic disentanglement indicated by the blue arrows.}
	\label{illus}
\end{figure}

However, despite the practical success of deep neural networks, such high-level representations are not as useful as one might hope. In the absence of constraints to enforce semantic meaning, deep neural networks will generate representations that are well-aligned with the statistics within the data, but at the expense of weakened semantic interpretability. So far, most successful image generation methods rely on annotated inputs~\cite{susskind2008generating,gauthier2014conditional}, or hand crafted steps, such as detection of facial or other landmarks~\cite{fan2015photo}. As a result, the ability of deep learning algorithms to build these high-level representations is not fully exploited. This could be partly attributed to the fact that the high-level representations are typically constructed based on statistics within the data, which, in general, does not align well with the semantics of the application~\cite{goodfellow2016deep,chen2016infogan}. In the illustrative example of Fig. \ref{illus} which depicts the latent space of a facial image dataset, the \textit{blonde} and \textit{blue-eye} attributes may be highly correlated statistically, leading to a representation with the principle direction of \textit{blue-eyed blonde}, rather than two separate and independent attributes. This latter attribute representation is probably more compact from an information theoretic point of view and it may be desirable in some applications such as compression or generation of fake images. However, it may not align well with other applications where attributes are required to be independent based on the application semantics. For example, in generation of an avatar, user may want to have independent control over eye-color and hair-color regardless of any strong correlation between these two attributes in the application domain.

One way to overcome this is to use a dataset with annotations of the semantic attributes and approach the problem as a supervised learning problem~\cite{liu2015faceattributes}. However, especially with large amounts of data, this is an expensive and error-prone task, where annotation difficulty increases with subtlety and number of semantic attributes.

Numerous models of real-world phenomenon are available in most domains. For example, in the domain of computer vision, many realistic 3D object models and rendering software have been developed ~\cite{makehuman,chang2015shapenet}, where semantic attributes (e.g., eye color) can be explicitly controlled. Hence, it is possible to construct a large dataset that is fully and accurately annotated and utilize these attributes in a way that forms the backbone of the representation of the real dataset.

In this paper, a synthetic dataset of over 500,000 facial images is algorithmically constructed using a physical model with a set of desired semantic facial attributes. In the first training step, the synthetic dataset is used to create a semantically-disentangled encoder for the desired attributes. In the second training step, a dataset of real facial images is used to train a VAE-GAN hybrid network where the latent space is a concatenation of the synthetic encoder output and a real encoder output. Noise with a special distribution is used both for the encoder and decoder to combat overfitting and achieve the best domain adaptation and generation performance without knowing any of the real data labels. It is demonstrated that the proposed SYNTH-VAE-GAN network can achieve a semantically-disentangled set of attributes for the real image domain.


The contributions of this paper can be listed as below:

1) A domain adaptation methodology that produces a \textit{semantically-disentangled} representation and associated predictive-generative model without relying on any labels for real data: Semantically-disentangled attributes are defined and enforced through an automatically-labeled synthetic dataset. An encoder for the synthetic dataset is trained and used to augment the real data encoder. It is demonstrated that SYNTH-VAE-GAN leads to a substantial disentanglement that is a 5-fold contrast between attributes.

2) \textit{Statistical whitening and decoupling} of the synthetic data parameters for creating a robust model manifold with a desired set of attributes: Model data parameters are sampled independently and based on a maximum entropy distribution to effectively form a statistical basis for the model manifold. This means that certain parameter combinations are allowed, even if they are not statistically plausible (female with beard or mustache as an example). It is shown that SYNTH-VAE-GAN leads to a 2-fold reduction in the average correlation between attributes in a test video coding example.

\section{Related Work}

Use of synthetic data to improve deep learning tasks have been studied by various researchers. In~\cite{tobin2017domain}, authors utilize simulated examples to facilitate classification and estimation tasks for achieving scene understanding. In~\cite{straka2010person,liu20163d}, authors use rendered faces to estimate head pose angles. Similarly in~\cite{achmed2010upper}, authors make use of rendered bodies for training a network to estimate upper body pose. In~\cite{rusu2016sim}, authors use progressive transfer learning approaches to make use of experiences of a robot from a simulated domain to real one. In~\cite{heimann2014real}, authors develop a system that can learn from unlabeled images to improve localization of objects for medical imaging application. \cite{pan2010survey} has an excellent survey of transfer learning methods. \cite{junlin2016deep,storkey2009training} describe transfer learning methods that involve multiple domains. In \cite{gonzalez2018image}, authors demonstrate learning of disentangled representations using shared representations among multiple domains. While the approach in this work would definitely lead to disentanglement between shared and unique representations, it does not explicitly enforce disentanglement within the shared attributes. Moreover, it also does not explicitly constrain the representations to take on semantic meanings. Our goal in this paper is to demonstrate semantic disentanglement between each one of the attributes, which is different from that in \cite{gonzalez2018image}.

None of the above approaches utilize synthetic data for generation tasks. Probably one of the closest work to ours is in \cite{shrivastava2017learning}, where authors combined unlabeled real, and labeled synthetic images for generation of realistic eye images. Even though the end goal of achieving realistic eye images is attained, the resulting network is a transformative one that maps from synthetic images to real ones. In the method proposed in this paper (SYNTH-VAE-GAN), the goal is to achieve a conditional predictive-generative model without having to rely on a labeled real dataset which is different from \cite{shrivastava2017learning}.

\section{Method}

\begin{figure*}[t]
\centering
\includegraphics[width=\linewidth,height=3in]{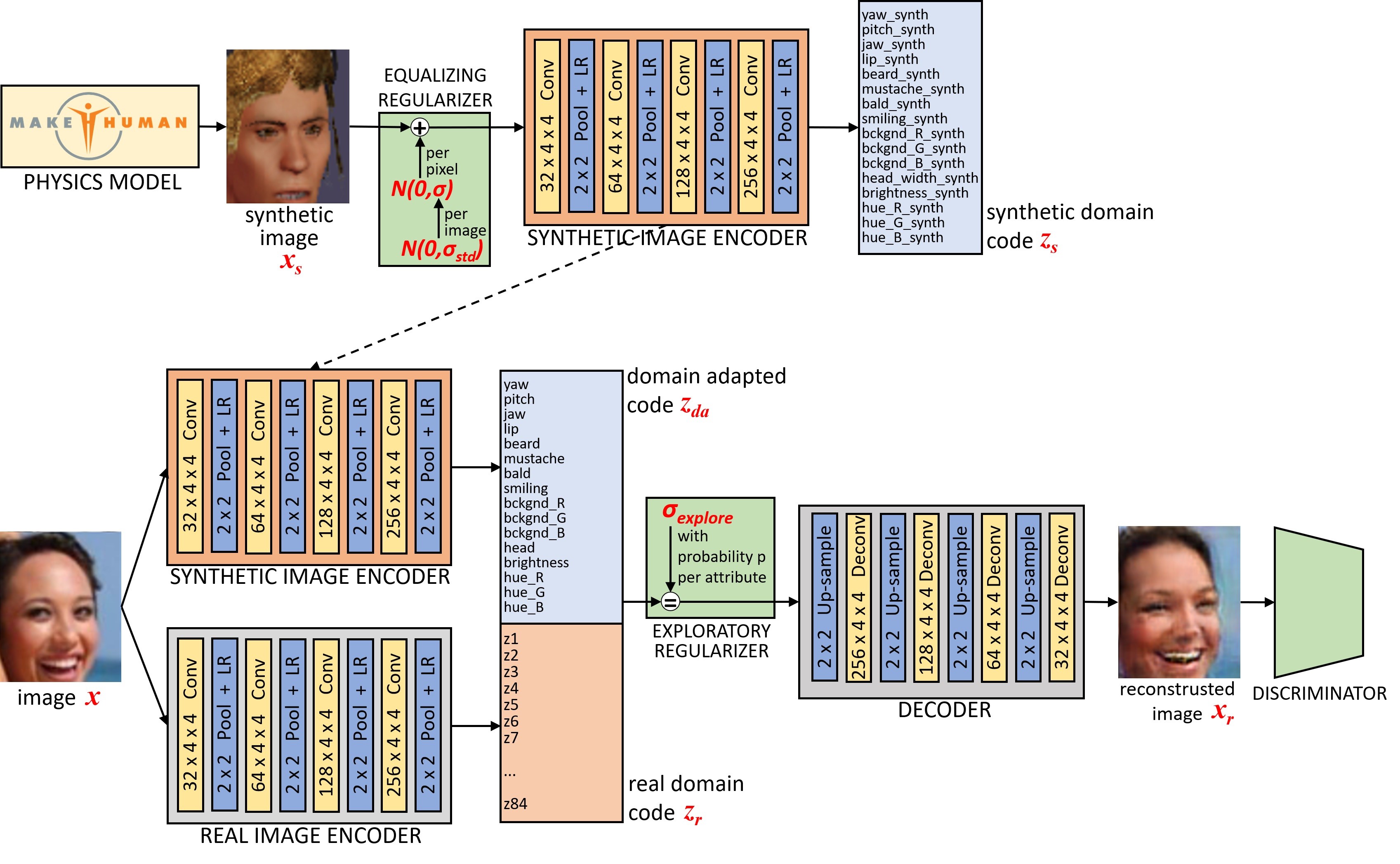}
\caption{SYNTH-VAE-GAN: A physics model is used to generate a synthetic dataset, which is subsequently used in training of a synthetic image encoder with semantically disentangled attributes $z_s$. The synthetic encoder is then augmented with a real image encoder and used in training of a VAE-GAN predictive-generative network for a real dataset. The resulting SYNTH-VAE-GAN network exhibits semantically disentangled attributes $z_{da}$ for the real image domain.}
\label{realnet}
\end{figure*}

In order to build the desired disentangled representation, the two-step procedure as illustrated in Fig.~\ref{realnet} is used:
\begin{enumerate}
    \item  A whitened and decoupled synthetic image dataset is generated using a synthetic model. The generated synthetic dataset is used to train a synthetic image encoder that produces disentangled and whitened attributes (see Fig. \ref{realnet} top).
    \item  A predictive-generative network is trained for the real data domain, but with the trained synthetic image encoder used to augment the real image domain attributes (see Fig. \ref{realnet} bottom).
\end{enumerate}

A detailed explanation of these steps is provided in the sections below.

\subsection{Construction of Disentangled Representations}

To achieve an accurate disentangled manifold, the dataset that is used in the training of the synthetic encoder is designed carefully to satisfy the following two conditions: 

\noindent \textbf{Whitened:} Each attribute is sampled using a maximum entropy distribution within the attribute's range. For attributes with finite mean and standard deviation in this paper, this is a uniform distribution for finite attribute range (i.e., $z_{s}(i) \sim U, \forall i$). The aim of whitening is two-fold: to balance the learning of statistically weak and strong attributes, and to ensure that all attributes are within prescribed and known ranges;

\noindent \textbf{Independent:} Each attribute is sampled independently from each other regardless of their statistics in real data (i.e., $z_{s}(i) \bot z_{s}(j), \forall i\neq j$). This means that certain attribute combination that is not plausible will be sampled as well as plausible combinations (such as a woman with a mustache).

After each full set of attributes $z_{s}$ is determined, a synthetic model $f$ is applied to create the corresponding image $x = f(z)$. In this paper, we use MakeHuman~\cite{makehuman}, an open source graphical modeling software, as the synthetic model $f$; Section~\ref{sec:exp} details this process. Random sampling is repeated to construct the synthetic dataset. This ensures that given a random sample $x$ from the synthetic dataset, the posterior distribution of $z$ is whitened with decoupled attributes, hence resulting in a disentangled learned representation. As shown in Fig.~\ref{realnet} top, a convolutional neural network (CNN) encoder is used to capture the disentangled representation of the synthetic image domain.

\subsection{Adaptation to the Real Domain}

Owning to the generalization that was used (to be detailed in Section~\ref{subsec:eqreg}), the encoder that is created in the previous step can be directly used as the predictor of a VAE~\cite{kingma2013auto} for the real image domain to create a predictor-generator pair. However, such direct use would limit the attributes to those from the synthetic domain, and all other attributes that are specific to the real domain would be ignored. To use the disentangled representation from the synthetic domain while allowing learning of new attributes from the real domain, a \textit{concatenated} latent space is utilized, as shown in Fig.~\ref{realnet} bottom. A separate, real image encoder is trained with the real dataset while fixing the synthetic image encoder, and the corresponding code $z_r$ is concatenated with the code from the synthetic image encoder $z_{da}$. Decoder uses the combination of disentangled and real domain attributes to reconstruct the original real image $x$. 

It is important to note here that the synthetic domain code $z_{da}$ is not guaranteed to be disentangled from the real domain code $z_r$ or within itself. As a result, to maximize the benefit of disentanglement, $z_{da}$ needs to cover most of the significant attributes (brightness, etc.), while $z_{r}$ is reserved for more subtle attributes (eye color, etc.). It has been observed that use of only a small number of attributes in $z_{da}$, or removing the generalization step (explained below in Section~\ref{subsec:eqreg}) leads to significant degradation in the quality of disentanglement.

\subsection{Equalizing Regularization}
\label{subsec:eqreg}

There are significant differences between the synthetic and real image domains and hence the encoder that is trained with the synthetic dataset does not generalize at all to the real image domain without employing a fine-tuned generalization approach. To achieve the desired generalization, an \textit{equalizing regularizer} is used: the regularization is achieved by injecting a noise $n$ with Normal distribution:
$\sigma \sim N(0, \sigma_{std})$ and $n \sim N(0, \sigma)$.
Here, $\sigma$ is randomly drawn once per image, $n$ is randomly drawn once per pixel, $N(\mu, \sigma )$ is the Normal distribution with a mean $\mu$ and standard deviation $\sigma$, and $\sigma_{std}$ is the standard deviation (per image) of the noise standard deviation (per pixel). This regularization ensures that some images are subject to only a small amount of noise, while others are subject to a large amount. This is found to be crucial for learning a large variety of attributes because subtle attributes (such as eye color) require the application of small amount of noise, while strong attributes (such as hair color) require a larger amount. It is noted here that the amount of generalization that needs to be applied, and quality of the resulting disentanglement depends on how similar synthetic and real image domain are. For example, it was observed that the quality of disentanglement significantly degraded if centering and size of faces in the synthetic and real domains were different. 

Quantification of disentanglement quality as a function of similarity between synthetic and real domains is a challenging task, and it is not addressed in this paper. For our current purposes, we evaluate the disentangled representations under the most favorable conditions, in which a physics-based model, i.e., MakeHuman~\cite{makehuman}, is available and acts as a close approximation to the real image domain. Intuitively, if the synthetic model is not beneficial in this situation, then it will not be beneficial in other conditions. Thus, it makes sense to start with this situation. Future work will need to consider other situations.

\subsection{Exploratory Regularization}

When the synthetic data image encoder is applied on the real image domain, the resulting code $z_{da}$ does not follow the prescribed ranges of attributes $z_{s}$. As a result, the code cannot be used for generative purposes (say, by drawing the code samples $z_{da}$) over the entire range. The reason is that any volume in the code space that is not visited by the decoder during training leads to images with deformities. To address this, a novel \textit{exploratory} regularizer inspired by reinforcement learning is used: with a probability $p = 0.01$, all $\sigma(l)$ values of the code $z_{da}$ are replaced by a value of $\sigma_{explore} = 2$ (see Eq. \ref{lossedge} for the definition of $\sigma(l)$).

\subsection{Training Details}

In training of the synthetic image encoder (Fig.~\ref{realnet} top), a standard regression approach with a L1-distance based loss is used (will not be detailed here). The encoder-decoder loss $\mathcal{L}_{\mathcal{G}}$ of the hybrid network (Fig.~\ref{realnet} bottom) is given as: $\mathcal{L}_{\mathcal{G}} = L_{r} + L_{l} + L_{g}$, where $L_{r}$ is the reconstruction loss; $L_{l}$ is the latent space loss; and $L_{g}$ is the generator loss. The individual losses in the above equation are given as:
\begin{eqnarray}
L_{r} & = & \frac{\alpha}{N}\underset{i,j,k}{\sum}\left|x_{i,j,k} - r_{i,j,k}\right| \enspace, \\
L_{l} & = & \beta \underset{l}{\sum}\left( (m_k)^2 + \sigma(l) - 1 - \log(\sigma)\right) \enspace, \\
L_{g} & = & - \gamma \log\left(\mathbb{D}(r_{i,j,k}) + \epsilon\right) \enspace,
\label{lossedge}
\end{eqnarray}
where $x_{i,j,k}$ is the input image value at pixel coordinates $i, j$ and channel $k$; $N$ is the number of pixels; $r_{i,j,k}$ is the reconstructed (output) image; $m$ is the output of the encoder that represents latent variable mean, $\sigma$ is the output of the encoder that represents latent variable standard deviation; $l$ is the attribute index; $\alpha$, $\beta$ and $\gamma$ are scaling constants that weigh the loss terms during optimization; $\mathbb{D}$ is the discriminator network functional which yields a probability $\mathcal{P}_{real}=\mathbb{D}(\cdot)$ (therefore, $\mathcal{P}_{fake} = 1-\mathbb{D}(\cdot)$) as output; and $\epsilon = 10^{-8}$ is a small number included for numerical stability purposes.

The discriminator loss $\mathcal{L}_{\mathcal{D}}$ is given as:
\begin{eqnarray}
\mathcal{L}_{\mathcal{D}} & = & - \log\left(1-\mathbb{D}(r_{i,j,k}) + \epsilon\right) \nonumber\\
 & & - \log\left(\mathbb{D}\left(x_{i,j,k}\right) + \epsilon\right) \enspace.
\end{eqnarray}

The reconstruction loss $L_{r}$ and latent space loss $L_{l}$ are the same that are typically used for VAE, while the discriminator loss $\mathcal{L}_{d}$ and generator loss $L_{g}$ are the same kind of loss typically employed in a GAN. Combination of these losses result in a VAE-GAN hybrid that can serve as both a predictor and a generator.

Batch size of $64$ and ADAM optimizer with initial learning rate of $0.0005$ are chosen for training. Learning rate is reduced by a factor of $4$ twice at approximately $1/3$ and $2/3$ points in the training. The loss weighting coefficients of $\alpha=1$, $\beta=8$ and $\gamma=0.03$ are used. Gradient clipping is used to help with stability.

Regularization for encoder and discriminator is provided by application of pixel-wise normally distributed input noise with a magnitude of $0.2$. Input noise regularization is favored to the popular Dropout method \cite{srivastava2014dropout}, since the latter seems to be more difficult to balance out with changing network parameters.

\section{Experiments}
\label{sec:exp}

\subsection{Real Dataset}
\begin{figure}[t]
	\begin{subfigure}{1.6in}
	\centering
	\includegraphics[scale = 0.24]{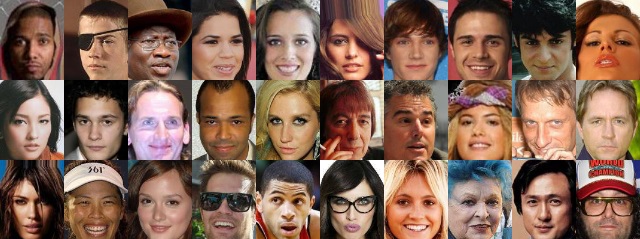}
	\caption{Celeb-A Dataset}
	\label{celeba}
	\end{subfigure}
	\begin{subfigure}{1.6in}
	\includegraphics[scale = 0.24]{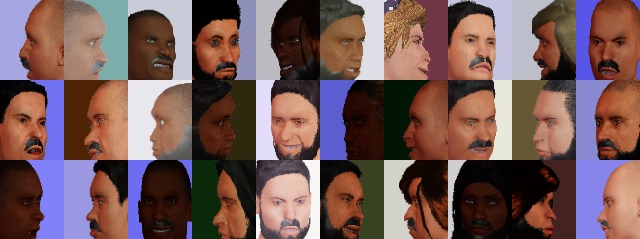}
	\caption{Synthetic Dataset}
	\label{caleba}
	\end{subfigure}

\caption{Examples of Celeb-A dataset (a), alongside examples from the synthetic dataset (b).}
\label{faceedgeex}
\end{figure}

The celebrity facial image dataset Celeb-A \cite{liu2015faceattributes} consisting of 202,599 images was used as the real dataset. The cropped and aligned version of the dataset was selected. Since our goal is to demonstrate representation learning from a synthetic dataset, a relatively small image size of 80 $\times$ 64 is used. Even though there are labels in this dataset, they were used only as reference and unobservable during training. Figure \ref{celeba} shows examples of the dataset.

\subsection{Synthetic Dataset}
In order to construct the labeled synthetic dataset of human faces, MakeHuman was used \cite{makehuman}. This tool allows for modeling of human face and body as well as clothing, accessories and facial and body poses with action units explicitly controllable. In this work, it has been utilized in scripted mode to generate a labeled dataset consisting of 500,000 images (Fig. \ref{caleba}). It is noted that the size of the synthetic dataset was determined by gradually increasing the size of the dataset until disentanglement of subtle features such as smile is achieved qualitatively.

The following short codes are used in Figs. \ref{entangled} and \ref{videos}: yaw (y), pitch (p), jaw (j), lip (l), age (a), beard (be), mustache (m), bald (ba), smiling (sm), background-R (bR), background-G (bG), background-B (bB), head width (hs), brightness (br), hue-R (hR), hue-G (hG) and hue-B (hB).

\subsection{Reference Models}

To serve as reference, a traditional unconditional (UC-VAE-GAN) and a conditional (C-VAE-GAN) model were trained based on the same VAE-GAN hybrid based network topology in Fig. \ref{realnet} but without the synthetic encoder. UC-VAE-GAN was trained in unsupervised fashion with a code the same size with that of SYNTH-VAE-GAN for fair comparison. A full set of labels that were used in supervised training of C-VAE-GAN but with additional (unsupervised) latent variables used to bring the size of the code to be the same as SYNTH-VAE-GAN, again for a fair comparison.

\subsection{Disentanglement Study --- Single Attribute}

In order to verify the resulting disentanglement, a separate test dataset with pairs of human face photos, where each pair consists of the same person with a smiling and non-smiling poses, are used.

\begin{figure}[t]
\centering
\includegraphics[width=\linewidth,height=2.0in]{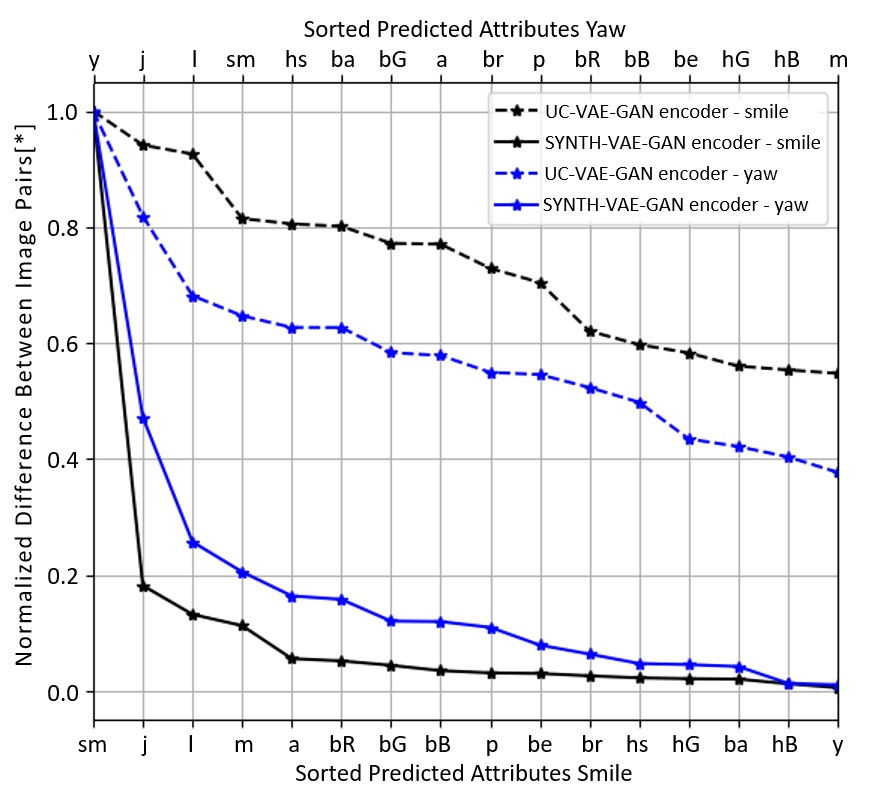}
\caption{Normalized difference between image pairs for smile/yaw attributes in encoder outputs. The attributes are sorted in decreasing order for ease of comparison. SYNTH-VAE-GAN encoder exhibits change mostly in a single attribute indicating improved disentanglement.}
\label{entangled}
\end{figure}

In order to measure disentanglement, first a real \textit{smile} attribute vector is estimated by taking the difference between code for smiling $z_{da}^s$ and non-smiling $z_{da}^{ns}$ images of the same pair of face images for each attribute $l$. Next the difference is normalized by the standard deviation of the code for the real dataset $\sigma_z$. Finally, this difference is normalized to yield a value of 1 for the smile attribute per pair to calculate normalized difference between image pairs $\Delta'$ and facilitate visualization of disentanglement:
\begin{align}
\Delta(l) = \frac{1}{\sigma_z (l) N_P}\underset{P}{\sum} (z_{da}^s (l) - z_{da}^{ns} (l)) \enspace, \\
\Delta'(l) = \frac{\Delta(l)}{\underset{l}{max}(\Delta(l))} \enspace.
\end{align}
where $P$ is the set of pairs, and $N_P$ is the number of pairs. It is expected that a perfect disentanglement would lead to only one attribute from encoder outputs being found to be strongly correlated to the estimated smile vector. Figure~\ref{entangled} shows the normalized difference between image pairs for the smile and yaw attributes, and for the UC-VAE-GAN and SYNTH-VAE-GAN encoders.

\begin{figure*}[t]
\centering
\includegraphics[width=6.8 in]{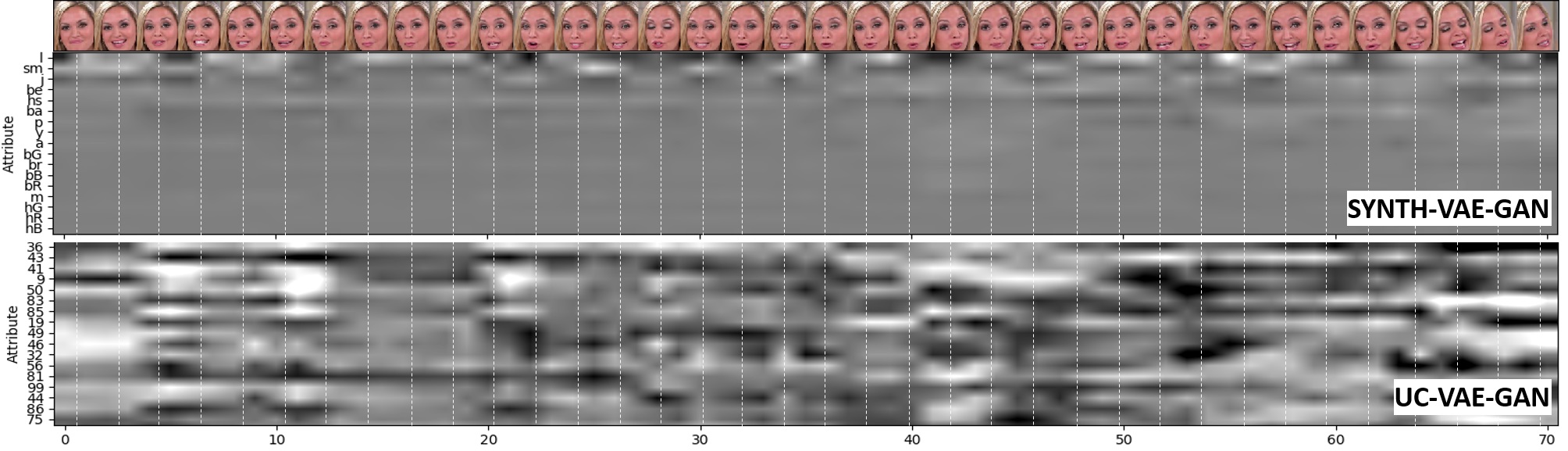}
\caption{A brightness image depicting the attribute values as a function of frame index for video of a human speaker. The brightness value range is from -5 (black) to 5 (white). SYNTH-VAE-GAN encoder results in only 2-3 significantly varying attributes indicating disentanglement among attributes.}
\label{videos}
\end{figure*}

It can be seen in Fig. \ref{entangled} that multiple UC-VAE-GAN encoder values are found to be approximately equally correlated to the estimated smile vector, while SYNTH-VAE-GAN encoder exhibited large change in only one attribute. This demonstrates that SYNTH-VAE-GAN encoder indeed can successfully semantically decouple attributes. It is noted, however, that the decoupling is not perfect: Some weak changes are also observed in jaw, mustache and brightnesses with magnitudes less than 1/5'th of the smile attribute. This is attributed mostly to errors in generalization of the synthetic dataset to the real dataset.

\subsection{Disentanglement Study --- Video Sequence}

In order to study the disentanglement effect for a wider range of attributes, a test video of a human speaker is used. The video is cropped and centered to the face of the speaker to match the encoder dataset. A total of 71 frames are used. Figure \ref{videos} shows an image constructed from $\Delta'(l)$ plotted as a function of frame index for both UC-VAE-GAN and SYNTH-VAE-GAN encoders. 

It can be seen in Fig. \ref{videos} that the largest 16 attributes for the UC-VAE-GAN encoder show equally large variations, while the SYNTH-VAE-GAN encoder shows large variations on only 2-3 of the attributes, namely: lip, smile and jaw. This indicates that, in the SYNTH-VAE-GAN encoder, representation of the speaking person is distributed to less attributes and it is better disentangled compared to the UC-VAE-GAN encoder.

To quantify the disentanglement, correlation coefficient of different attributes during the person's speech is tabulated in Table \ref{corrreal}. It is noted that disentanglement does not necessarily lead to uncorrelated attributes, however magnitude of correlation increases by entanglement and hence correlation can be used as a rough measure of entanglement. 

\begin{table}[t]
\centering
\caption{UC-VAE-GAN and SYNTH-VAE-GAN Encoder Correlations}
\label{corrreal}
\includegraphics[width=3.25 in,height=1in]{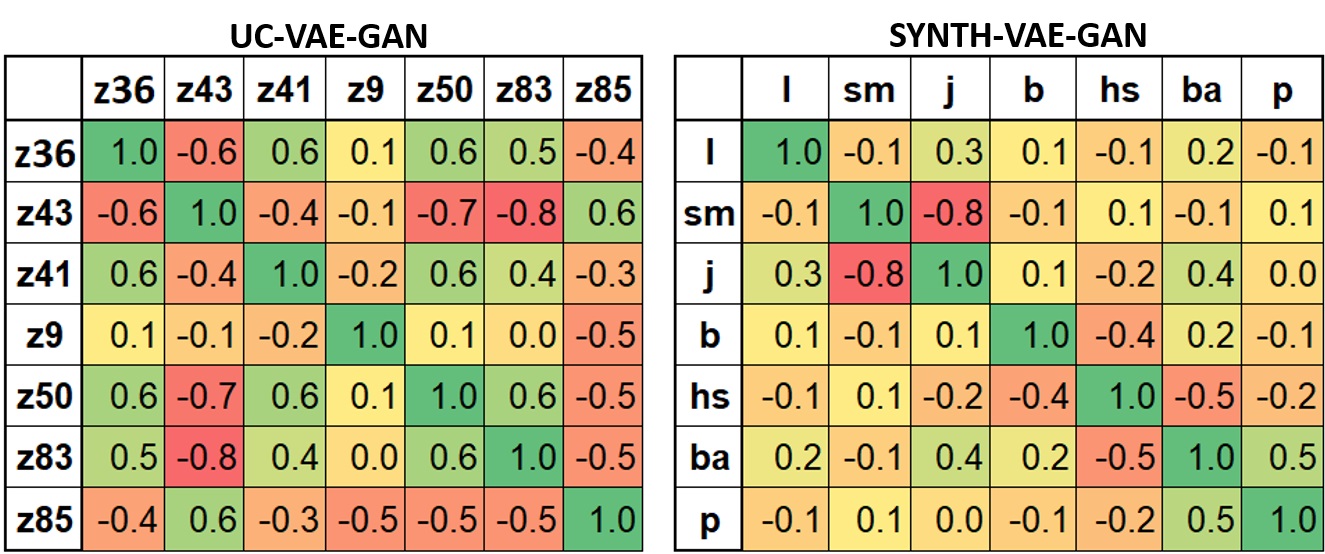}
\end{table}

\begin{figure*}[ht]
\centering
	\begin{subfigure}{2.25in}
    \includegraphics[width=2.15 in, trim=0 0 0 -80mm]{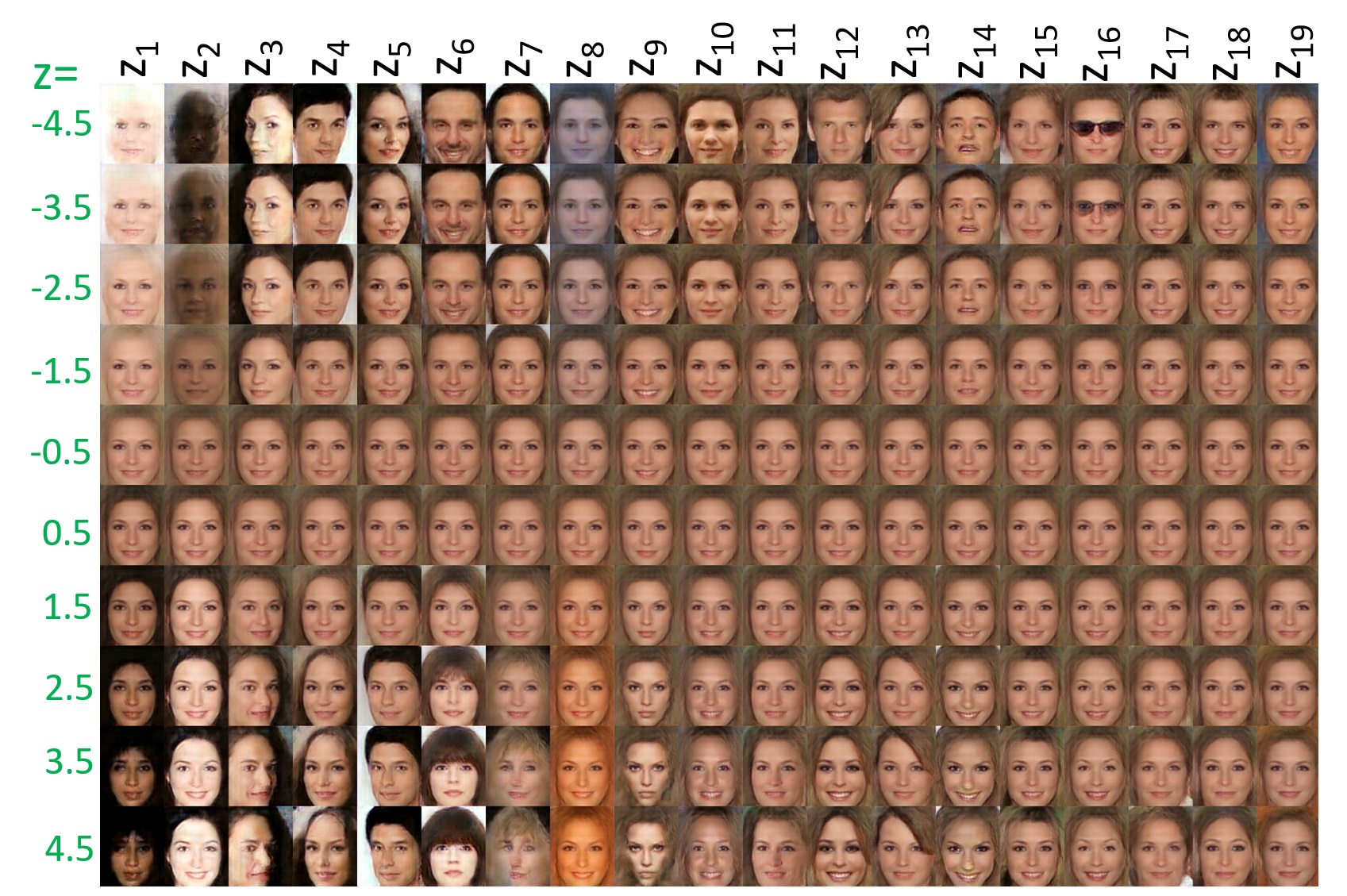}
    \caption{UC-VAE-GAN}
    \label{uncondlatent}
	\end{subfigure}
	\begin{subfigure}{2.25in}
    \centering
    \includegraphics[width=2.25 in, trim=0 0 0 -22mm]{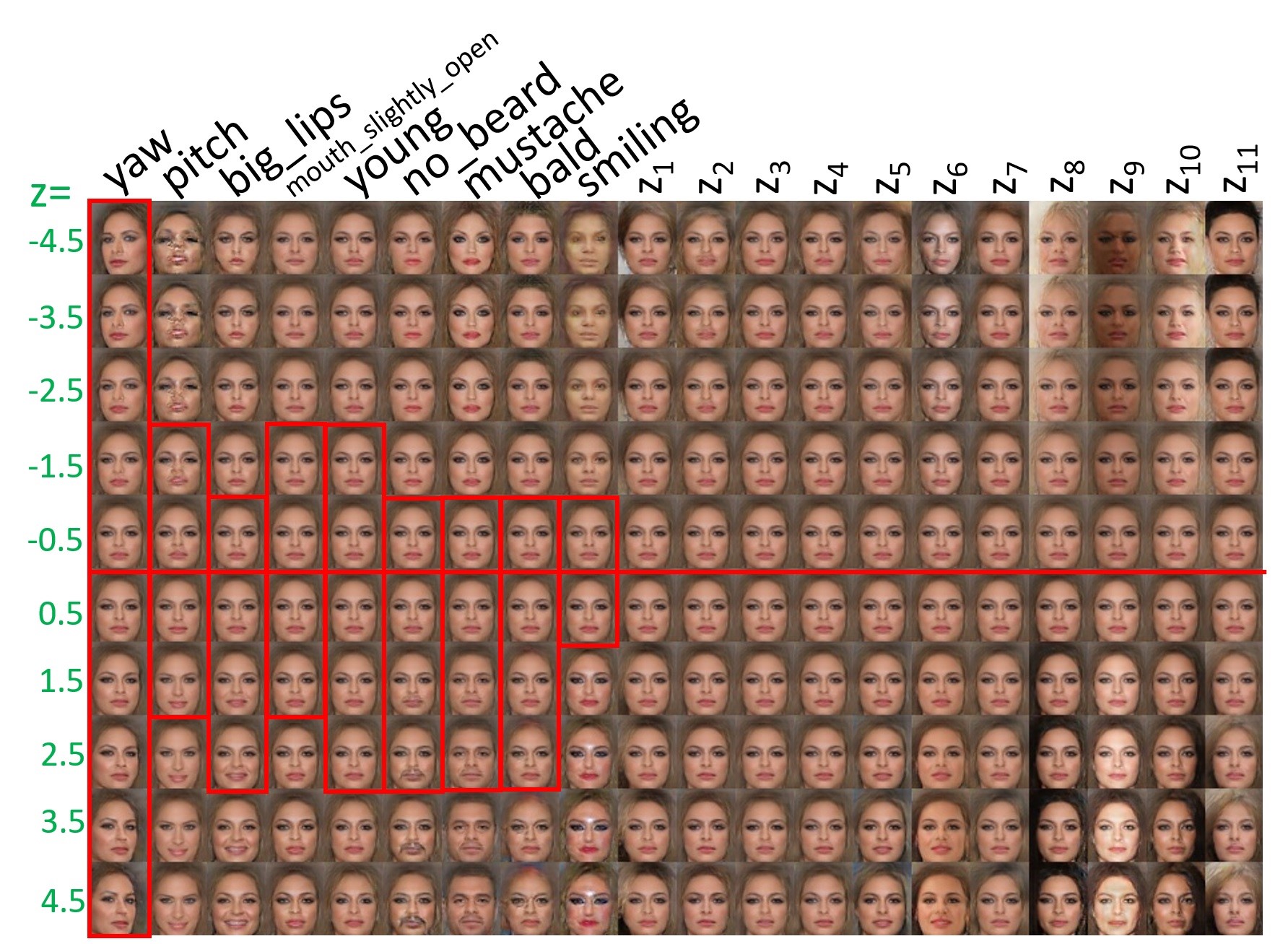}
    \caption{C-VAE-GAN}
    \label{condlatent}
	\end{subfigure}
	\begin{subfigure}{2.25in}
    \centering
    \includegraphics[width=2.25 in]{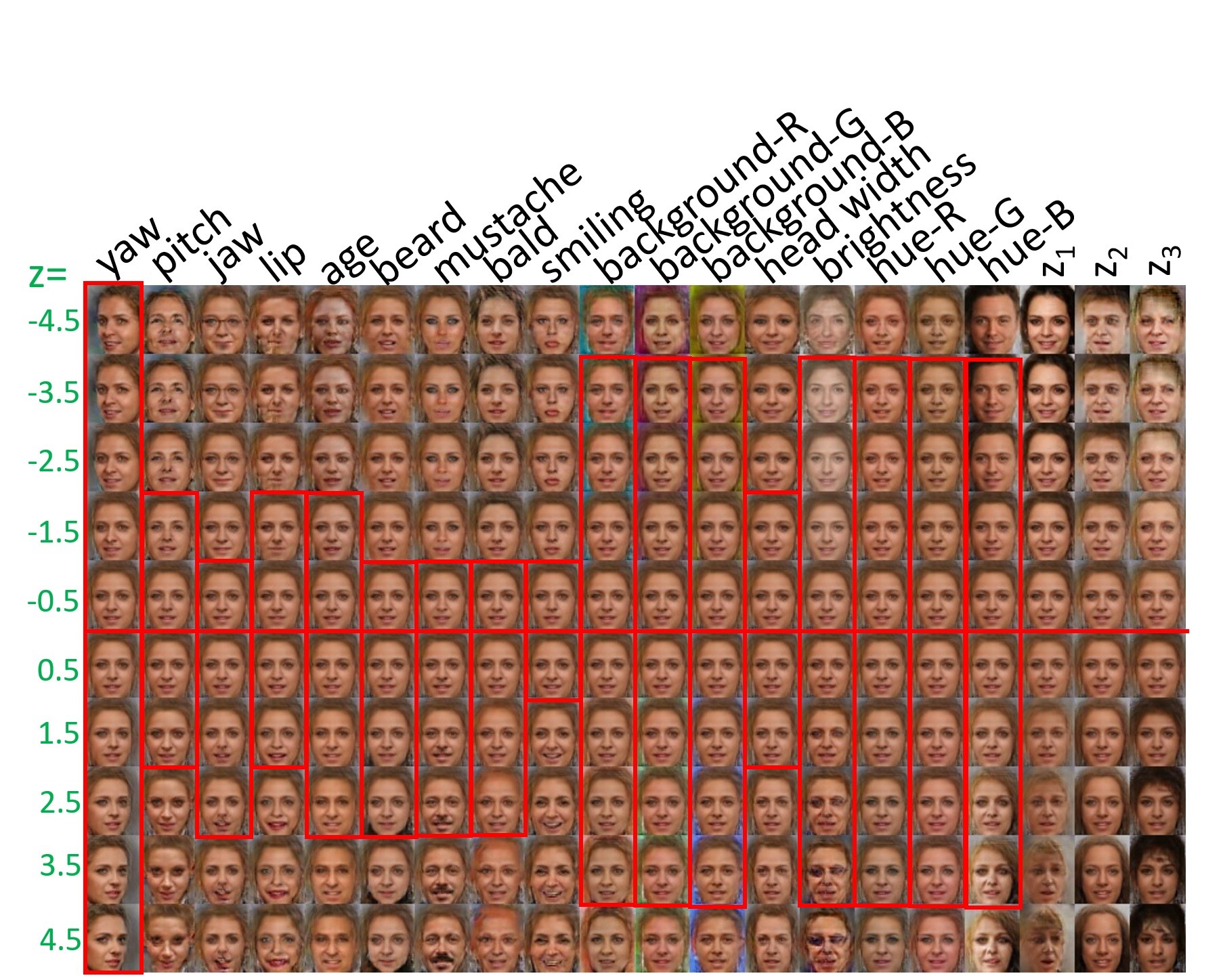}
    \caption{SYNTH-VAE-GAN}
    \label{tflatent}
	\end{subfigure}
\caption{Visualization of the latent space along each variable dimension for (a) the UC-VAE-GAN ; (b) C-VAE-GAN; and (c) SYNTH-VAE-GAN from Fig.~\ref{realnet}. Red boxes indicate labeled attribute ranges that are used during training. Please zoom and view it on a computer screen.}
\label{latent}
\end{figure*}

It can be seen in Table \ref{corrreal} that the most significant 7 attributes of the SYNTH-VAE-GAN encoder are much less correlated than that of the UC-VAE-GAN encoder. The average absolute correlation coefficient of the SYNTH-VAE-GAN encoder is calculated to be $0.22$, while it is $0.43$ for C-VAE-GAN, which is a 2-fold reduction in average correlation for SYNTH-VAE-GAN.

\begin{figure*}[t]
	\begin{subfigure}{1.68 in}
    \centering
    \includegraphics[width=1.5 in]{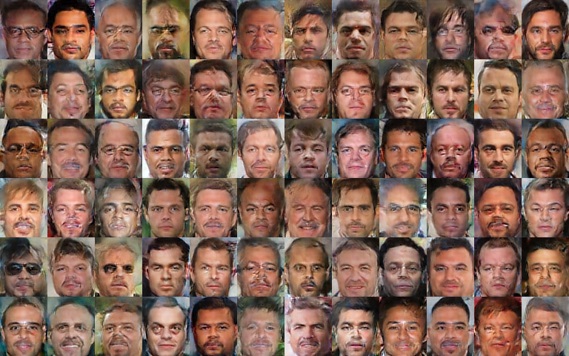} \caption{mustache | C-}
    \label{gen_mustache_real}
	\end{subfigure}
	\begin{subfigure}{1.68 in}
    \centering
    \includegraphics[width=1.5 in]{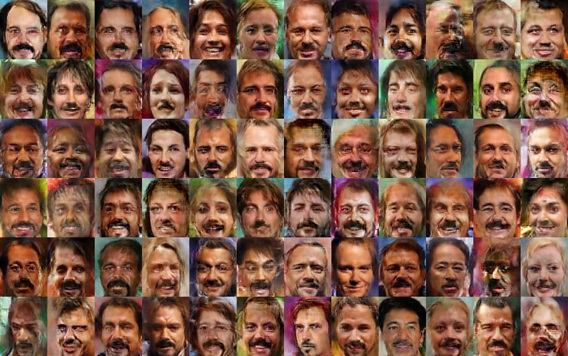} \caption{mustache | SYNTH-}
    \label{gen_mustache_trans}
	\end{subfigure}
	\begin{subfigure}{1.68 in}
    \centering
    \includegraphics[width=1.5 in]{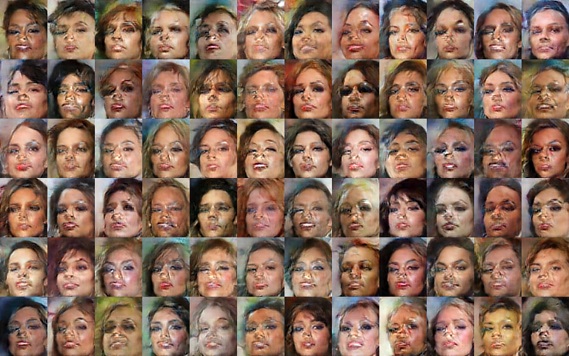} \caption{pitch, yaw | C-}
    \label{gen_pitchyaw_real}
	\end{subfigure}
	\begin{subfigure}{1.68 in}
    \centering
    \includegraphics[width=1.5 in]{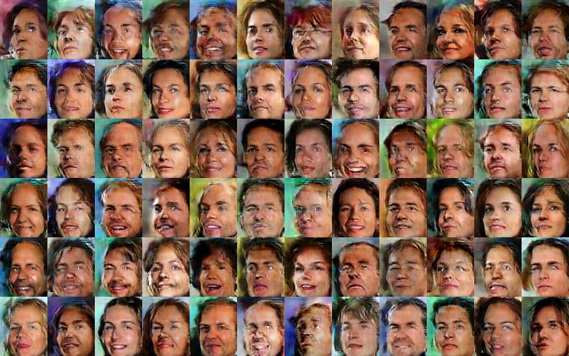} \caption{pitch, yaw | SYNTH-}
    \label{gen_pitchyaw_trans}
	\end{subfigure}

\caption{Conditionally generated fake examples of celebrities with mustaches and specific head pitch and yaw angles. SYNTH-VAE-GAN can produce images with pronounced mustache types.}
\label{gen}
\end{figure*}

\begin{figure*}[t]
    \centering
	\begin{subfigure}{2.25in}
    \centering
    \includegraphics[width=1.5 in]{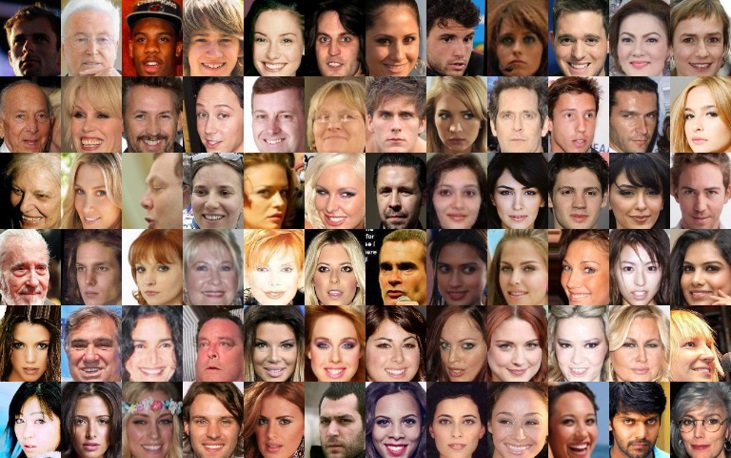} \caption{original}
    \label{trans_original}
	\end{subfigure}
	\begin{subfigure}{2.25in}
    \centering
    \includegraphics[width=1.5 in]{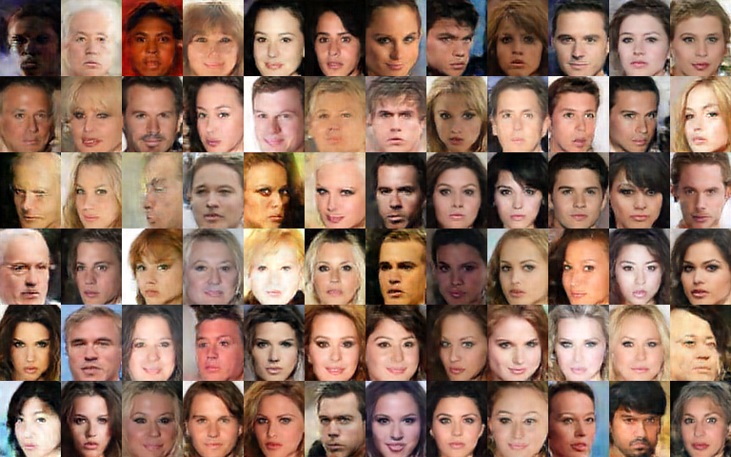} \caption{no smile}
    \label{trans_nosmile}
	\end{subfigure}
	\begin{subfigure}{2.25in}
    \centering
    \includegraphics[width=1.5 in]{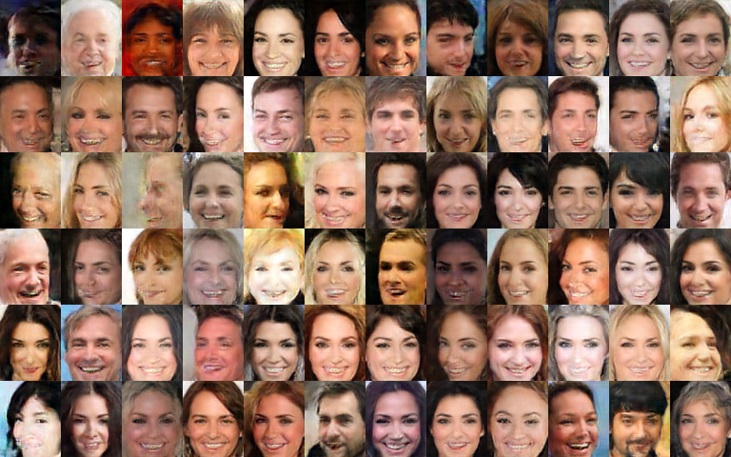} \caption{smile}
    \label{trans_smile}
	\end{subfigure}

\caption{Examples of transformative processing of images using SYNTH-VAE-GAN.}
\label{trans}
\end{figure*}

\subsection{Learned Latent Space}

Figure \ref{uncondlatent} shows a one-hot visualization of the latent space of UC-VAE-GAN at the image output $x_r$. This representation reflects the results with an unsupervised learning approach and reflects entirely the internal statistics of the data. The latent variables are ordered, from high to low, based on the $L_2$ norm of pixel-wise difference between the code values of $-5$ and $5$. Only the largest 19 latent variables are shown. It can be seen that a large number of important attributes are captured such as color of skin, background, head yaw and facial characteristics. However, most of the attributes of this representation are entangled: skin color and background color, facial characteristics and head yaw angle, and so on. As mentioned before, this may be undesirable in certain applications where a disentangled representation is required.

Figure~\ref{condlatent} shows the visualization of the latent space along each variable dimension for C-VAE-GAN. It can be seen that the use of conditional variables decouples attributes, with some undesirable side-effects due to dependence between the conditional variables. As an example, the attribute \textit{beard} seems to have adopted the image of a strong mustache with no beard, while attribute \textit{mustache} seems to have adopted a very faint mustache. Similarly, the attribute \textit{big lips} seems to have adopted an image of smile, while attribute \textit{smile} seems to have adopted a faint smile with a strong lip color. A solution to this could be to design independent labels, but this would not be feasible due to (a) the difficulty of creating or adjusting labels due to high cost of annotations; (b) there may not be sufficient samples for certain improbable combinations (such as a female with a mustache). 
Figure \ref{tflatent} shows the visualization of the latent space along each variable dimension for SYNTH-VAE-GAN. The first 16 latent space indices correspond to domain adapted attributes, while the rest are free latent variables. It can be seen from Fig. \ref{tflatent}, the manifold of domain adapted attributes variables has successfully been captured. In addition, free variables acquired representation of other parameters which were not domain adapted such as hair type or skin color.

\subsection{Generative Examples}

Figure \ref{gen} compares the generated fake examples of celebrities using C-VAE-GAN and SYNTH-VAE-GAN. The first plot, Fig. \ref{gen_mustache_trans}, shows SYNTH-VAE-GAN conditioned on the presence of mustaches, while the second plot, Fig. \ref{gen_mustache_real}, shows the C-VAE-GAN conditioned on the presence of mustaches. 

It can be seen from the comparison of the two sets of generated examples that SYNTH-VAE-GAN can generate images with the very specific kind of mustache which was used in the construction of the synthetic dataset ---a very pronounced mustache type. On the other hand, C-VAE-GAN produces mostly faint mustaches which are the most common type in the celebrity dataset. Hence, the proposed method was able to overcome the difficulty of generating a rare mustache type. It is noted that this is also a disadvantage since it will be mainly limited to the types of features defined by the synthetic dataset.

Figures \ref{gen_pitchyaw_trans} and \ref{gen_pitchyaw_real} show a similar comparison for generation of images with specific head pitch and yaw angles. Both SYNTH-VAE-GAN and C-VAE-GAN show similar characteristics with minor differences.

\subsection{Transformative Examples}

Since SYNTH-VAE-GAN allows training of both an encoder and a corresponding decoder, it is possible to perform \textit{transformative} processing of images where: (i) a real image is transformed into code using the encoder; (ii) a modification is made on the code on one of the desired attributes per latent space that was assembled; (iii) modified code is transformed back into an image using the decoder. Figure~\ref{trans} shows original images and two sets of transformed images where smile code is set to true or false as indicated. It can be seen via visual inspection that the original image is successfully modified to produce images with  \textit{no smile} and \textit{smile} attributes.

\section{Conclusion}

A domain adaptation method based on attribute representation transfer is used to build a conditional predictive-generative model without relying on any real data labels. This is accomplished by first, building a synthetic dataset with whitened and independent attributes as well as corresponding labels; second, training an encoder for the synthetic dataset; and finally, training a predictive-generative network using a latent variable that consists of a concatenation of the synthetic and real dataset encoders. It is demonstrated that the proposed method (SYNTH-VAE-GAN) can result in a 5-fold improvement in disentanglement, and a 2-fold improvement of average attribute correlation for the resulting representations, when compared to a unsupervised VAE-GAN. It can be used in both generative and transformative tasks as a replacement for the annotated real data, resulting in significant savings in annotation cost.

\noindent \textbf{Limitations.} \quad The proposed method assumes availability of a physics model with a sufficiently large set of attributes for generation of the synthetic dataset. In addition, the similarity between the synthetic domain and the real domain, and the number of attributes used in the physics model are crucial parameters to success of the method but they are difficult to quantify.

\noindent \textbf{Future work.} \quad Methods to quantify similarity between the synthetic and real domains can be investigated. Application can be extended to one of the many practical domains where labeled datasets are difficult to come-by. 

\bibliographystyle{aaai} \bibliography{my_ref}

\begin{thebibliography}{}

\bibitem[\protect\citeauthoryear{Achmed and Connan}{2010}]{achmed2010upper}
Achmed, I., and Connan, J.
\newblock 2010.
\newblock Upper body pose estimation towards the translation of south african
  sign language.
\newblock In {\em Proceedings of the Southern Africa Telecommunication Networks
  and Applications Conference},  427--432.

\bibitem[\protect\citeauthoryear{Chang \bgroup et al\mbox.\egroup
  }{2015}]{chang2015shapenet}
Chang, A.~X.; Funkhouser, T.~A.; Guibas, L.~J.; Hanrahan, P.; Huang, Q.-X.; Li,
  Z.; Savarese, S.; Savva, M.; Song, S.; Su, H.; Xiao, J.; Yi, L.; and Yu, F.
\newblock 2015.
\newblock Shapenet: An information-rich 3d model repository.
\newblock {\em CoRR} abs/1512.03012.

\bibitem[\protect\citeauthoryear{Chen \bgroup et al\mbox.\egroup
  }{2016}]{chen2016infogan}
Chen, X.; Duan, Y.; Houthooft, R.; Schulman, J.; Sutskever, I.; and Abbeel, P.
\newblock 2016.
\newblock Infogan: Interpretable representation learning by information
  maximizing generative adversarial nets.
\newblock In {\em Advances in Neural Information Processing Systems},
  2172--2180.

\bibitem[\protect\citeauthoryear{Fan \bgroup et al\mbox.\egroup
  }{2015}]{fan2015photo}
Fan, B.; Wang, L.; Soong, F.~K.; and Xie, L.
\newblock 2015.
\newblock Photo-real talking head with deep bidirectional lstm.
\newblock In {\em Acoustics, Speech and Signal Processing (ICASSP), 2015 IEEE
  International Conference on},  4884--4888.
\newblock IEEE.

\bibitem[\protect\citeauthoryear{Gauthier}{2014}]{gauthier2014conditional}
Gauthier, J.
\newblock 2014.
\newblock Conditional generative adversarial nets for convolutional face
  generation.
\newblock {\em Class Project for Stanford CS231N: Convolutional Neural Networks
  for Visual Recognition, Winter semester} 2014(5):2.

\bibitem[\protect\citeauthoryear{Gonzalez-Garcia, van~de Weijer, and
  Bengio}{2018}]{gonzalez2018image}
Gonzalez-Garcia, A.; van~de Weijer, J.; and Bengio, Y.
\newblock 2018.
\newblock Image-to-image translation for cross-domain disentanglement.
\newblock In {\em NeurIPS}.

\bibitem[\protect\citeauthoryear{Goodfellow \bgroup et al\mbox.\egroup
  }{2014}]{goodfellow2014generative}
Goodfellow, I.; Pouget-Abadie, J.; Mirza, M.; Xu, B.; Warde-Farley, D.; Ozair,
  S.; Courville, A.; and Bengio, Y.
\newblock 2014.
\newblock Generative adversarial nets.
\newblock In {\em Advances in neural information processing systems},
  2672--2680.

\bibitem[\protect\citeauthoryear{Goodfellow \bgroup et al\mbox.\egroup
  }{2016}]{goodfellow2016deep}
Goodfellow, I.; Bengio, Y.; Courville, A.; and Bengio, Y.
\newblock 2016.
\newblock {\em Deep learning}, volume~1.
\newblock MIT press Cambridge.

\bibitem[\protect\citeauthoryear{Heimann \bgroup et al\mbox.\egroup
  }{2014}]{heimann2014real}
Heimann, T.; Mountney, P.; John, M.; and Ionasec, R.
\newblock 2014.
\newblock Real-time ultrasound transducer localization in fluoroscopy images by
  transfer learning from synthetic training data.
\newblock {\em Medical image analysis} 18(8):1320--1328.

\bibitem[\protect\citeauthoryear{Ioffe and Szegedy}{2015}]{ioffe2015batch}
Ioffe, S., and Szegedy, C.
\newblock 2015.
\newblock Batch normalization: Accelerating deep network training by reducing
  internal covariate shift.
\newblock In {\em ICML}.

\bibitem[\protect\citeauthoryear{Junlin \bgroup et al\mbox.\egroup
  }{2016}]{junlin2016deep}
Junlin, H.; Jiwen, L.; Yap-Peng, T.; and Jie, Z.
\newblock 2016.
\newblock Deep transfer metric learning.
\newblock {\em IEEE transactions on image processing: a publication of the IEEE
  Signal Processing Society} 25(12):5576--5588.

\bibitem[\protect\citeauthoryear{Kingma and Welling}{2014}]{kingma2013auto}
Kingma, D.~P., and Welling, M.
\newblock 2014.
\newblock Auto-encoding variational bayes.
\newblock {\em CoRR} abs/1312.6114.

\bibitem[\protect\citeauthoryear{Krizhevsky, Sutskever, and
  Hinton}{2012}]{krizhevsky2012imagenet}
Krizhevsky, A.; Sutskever, I.; and Hinton, G.~E.
\newblock 2012.
\newblock Imagenet classification with deep convolutional neural networks.
\newblock In {\em Advances in neural information processing systems},
  1097--1105.

\bibitem[\protect\citeauthoryear{Liu \bgroup et al\mbox.\egroup
  }{2015}]{liu2015faceattributes}
Liu, Z.; Luo, P.; Wang, X.; and Tang, X.
\newblock 2015.
\newblock Deep learning face attributes in the wild.
\newblock In {\em Proceedings of International Conference on Computer Vision
  (ICCV)}.

\bibitem[\protect\citeauthoryear{Liu \bgroup et al\mbox.\egroup
  }{2016}]{liu20163d}
Liu, X.; Liang, W.; Wang, Y.; Li, S.; and Pei, M.
\newblock 2016.
\newblock 3d head pose estimation with convolutional neural network trained on
  synthetic images.
\newblock In {\em Image Processing (ICIP), 2016 IEEE International Conference
  on},  1289--1293.
\newblock IEEE.

\bibitem[\protect\citeauthoryear{MakeHumanCommunity}{2018}]{makehuman}
MakeHumanCommunity.
\newblock 2018.
\newblock {Make Human: Open source tool for making 3D characters}.

\bibitem[\protect\citeauthoryear{Pan, Yang, and others}{2010}]{pan2010survey}
Pan, S.~J.; Yang, Q.; et~al.
\newblock 2010.
\newblock A survey on transfer learning.
\newblock {\em IEEE Transactions on knowledge and data engineering}
  22(10):1345--1359.

\bibitem[\protect\citeauthoryear{Radford, Metz, and
  Chintala}{2016}]{radford2015unsupervised}
Radford, A.; Metz, L.; and Chintala, S.
\newblock 2016.
\newblock Unsupervised representation learning with deep convolutional
  generative adversarial networks.
\newblock {\em CoRR} abs/1511.06434.

\bibitem[\protect\citeauthoryear{Rusu \bgroup et al\mbox.\egroup
  }{2017}]{rusu2016sim}
Rusu, A.~A.; Vecerik, M.; Roth{\"o}rl, T.; Heess, N.; Pascanu, R.; and Hadsell,
  R.
\newblock 2017.
\newblock Sim-to-real robot learning from pixels with progressive nets.
\newblock In {\em CoRL}.

\bibitem[\protect\citeauthoryear{Sermanet \bgroup et al\mbox.\egroup
  }{2014}]{sermanet2013overfeat}
Sermanet, P.; Eigen, D.; Zhang, X.; Mathieu, M.; Fergus, R.; and LeCun, Y.
\newblock 2014.
\newblock Overfeat: Integrated recognition, localization and detection using
  convolutional networks.
\newblock In {\em International Conference on Learning Representations}.

\bibitem[\protect\citeauthoryear{Shrivastava \bgroup et al\mbox.\egroup
  }{2017}]{shrivastava2017learning}
Shrivastava, A.; Pfister, T.; Tuzel, O.; Susskind, J.; Wang, W.; and Webb, R.
\newblock 2017.
\newblock Learning from simulated and unsupervised images through adversarial
  training.
\newblock In {\em The IEEE Conference on Computer Vision and Pattern
  Recognition (CVPR)}, volume~3, ~6.

\bibitem[\protect\citeauthoryear{Simonyan and
  Zisserman}{2015}]{simonyan2014very}
Simonyan, K., and Zisserman, A.
\newblock 2015.
\newblock Very deep convolutional networks for large-scale image recognition.
\newblock In {\em International Conference on Learning Representations}.

\bibitem[\protect\citeauthoryear{Srivastava \bgroup et al\mbox.\egroup
  }{2014}]{srivastava2014dropout}
Srivastava, N.; Hinton, G.; Krizhevsky, A.; Sutskever, I.; and Salakhutdinov,
  R.
\newblock 2014.
\newblock Dropout: a simple way to prevent neural networks from overfitting.
\newblock {\em The Journal of Machine Learning Research} 15(1):1929--1958.

\bibitem[\protect\citeauthoryear{Storkey}{2009}]{storkey2009training}
Storkey, A.~J.
\newblock 2009.
\newblock When training and test sets are different: characterising learning
  transfer.
\newblock In {\em In Dataset Shift in Machine Learning},  3--28.
\newblock MIT Press.

\bibitem[\protect\citeauthoryear{Straka \bgroup et al\mbox.\egroup
  }{2010}]{straka2010person}
Straka, M.; Urschler, M.; Storer, M.; Bischof, H.; Birchbauer, J.~A.; and
  Center, S.~B.
\newblock 2010.
\newblock Person independent head pose estimation by non-linear regression and
  manifold embedding.
\newblock In {\em Proc. 34rd Workshop of the Austrian Association for Pattern
  Recognition (AAPR/OAGM)}.

\bibitem[\protect\citeauthoryear{Susskind \bgroup et al\mbox.\egroup
  }{2008}]{susskind2008generating}
Susskind, J.~M.; Hinton, G.~E.; Movellan, J.~R.; and Anderson, A.~K.
\newblock 2008.
\newblock Generating facial expressions with deep belief nets.
\newblock In {\em Affective Computing}. InTech.

\bibitem[\protect\citeauthoryear{Szegedy \bgroup et al\mbox.\egroup
  }{2017}]{szegedy2017inception}
Szegedy, C.; Ioffe, S.; Vanhoucke, V.; and Alemi, A.~A.
\newblock 2017.
\newblock Inception-v4, inception-resnet and the impact of residual connections
  on learning.
\newblock In {\em AAAI}, volume~4, ~12.

\bibitem[\protect\citeauthoryear{Tobin \bgroup et al\mbox.\egroup
  }{2017}]{tobin2017domain}
Tobin, J.; Fong, R.; Ray, A.; Schneider, J.; Zaremba, W.; and Abbeel, P.
\newblock 2017.
\newblock Domain randomization for transferring deep neural networks from
  simulation to the real world.
\newblock In {\em Intelligent Robots and Systems (IROS), 2017 IEEE/RSJ
  International Conference on},  23--30.
\newblock IEEE.

\bibitem[\protect\citeauthoryear{Wu \bgroup et al\mbox.\egroup
  }{2016}]{wu2016learning}
Wu, J.; Zhang, C.; Xue, T.; Freeman, B.; and Tenenbaum, J.
\newblock 2016.
\newblock Learning a probabilistic latent space of object shapes via 3d
  generative-adversarial modeling.
\newblock In {\em Advances in Neural Information Processing Systems},  82--90.

\bibitem[\protect\citeauthoryear{Yosinski \bgroup et al\mbox.\egroup
  }{2015}]{yosinski2015understanding}
Yosinski, J.; Clune, J.; Nguyen, A.~M.; Fuchs, T.~J.; and Lipson, H.
\newblock 2015.
\newblock Understanding neural networks through deep visualization.
\newblock {\em CoRR} abs/1506.06579.

\end{thebibliography}

\end{document}